\documentclass[sigconf, screen]{acmart}

\usepackage{multirow}
\usepackage{enumitem}
\usepackage{svg}
\usepackage{caption}
\usepackage{subcaption}
\usepackage{esvect}
\AtBeginDocument{%
  }

\copyrightyear{2024}
\acmYear{2024}
\setcopyright{rightsretained}
\acmConference[SIGSPATIAL '24]{The 32nd ACM International Conference on Advances in Geographic Information Systems}{October 29-November 1, 2024}{Atlanta, GA, USA}
\acmBooktitle{The 32nd ACM International Conference on Advances in Geographic Information Systems (SIGSPATIAL '24), October 29-November 1, 2024, Atlanta, GA, USA}
\acmDOI{10.1145/3678717.3691303}
\acmISBN{979-8-4007-1107-7/24/10}

\acmSubmissionID{216}

\begin{document}

\title{TrajGPT: Controlled Synthetic Trajectory Generation Using a Multitask Transformer-Based Spatiotemporal Model}

\author{Shang-Ling Hsu}
\affiliation{%
  \institution{University of Southern California}
  \city{Los Angeles}
  \state{California}
  \country{USA}
}
\email{hsushang@usc.edu}

\author{Emmanuel Tung}
\affiliation{%
  \institution{Novateur Research Solutions}
  \city{Ashburn}
  \state{Virginia}
  \country{USA}
}
\email{etung@novateur.ai}

\author{John Krumm}
\affiliation{%
  \institution{University of Southern California}
  \city{Los Angeles}
  \state{California}
  \country{USA}
}
\email{jkrumm@usc.edu}

\author{Cyrus Shahabi}
\affiliation{%
  \institution{University of Southern California}
  \city{Los Angeles}
  \state{California}
  \country{USA}
}
\email{shahabi@usc.edu}

\author{Khurram Shafique}
\affiliation{%
  \institution{Novateur Research Solutions}
  \city{Ashburn}
  \state{Virginia}
  \country{USA}
}
\email{kshafique@novateur.ai}

\renewcommand{\shortauthors}{Hsu et al.}

\begin{abstract}
Human mobility modeling from GPS-trajectories and synthetic trajectory generation are crucial for various applications, such as urban planning, disaster management and epidemiology. Both of these tasks often require filling gaps in a partially specified sequence of visits, -- a new problem that we call ``controlled'' synthetic trajectory generation. Existing methods for next-location prediction or synthetic trajectory generation cannot solve this problem as they lack the mechanisms needed to constrain the generated sequences of visits. Moreover, existing approaches (1) frequently treat space and time as independent factors, an assumption that fails to hold true in real-world scenarios, and (2) suffer from challenges in accuracy of temporal prediction as they fail to deal with mixed distributions and the inter-relationships of different modes with latent variables (e.g., day-of-the-week). These limitations become even more pronounced when the task involves filling gaps within sequences instead of solely predicting the next visit.
  
We introduce TrajGPT, a transformer-based, multi-task, joint spatiotemporal generative model to address these issues. Taking inspiration from large language models, TrajGPT poses the problem of controlled trajectory generation as that of text infilling in natural language. TrajGPT integrates the spatial and temporal models in a transformer architecture through a Bayesian probability model that ensures that the gaps in a visit sequence are filled in a spatiotemporally consistent manner. Our experiments on public and private datasets demonstrate that TrajGPT not only excels in controlled synthetic visit generation but also outperforms competing models in next-location prediction tasks--Relatively, TrajGPT achieves a 26-fold improvement in temporal accuracy while retaining more than 98\% of spatial accuracy on average.
\end{abstract}





\maketitle

\section{Introduction}
Modeling human mobility is important for understanding traffic, urban dynamics, commerce, health, and equity. Ideally, researchers and practitioners would have access to relevant, detailed visit sequences of large numbers of people. In reality, however, it is difficult to get a large volume of high-quality visit sequences, due to concerns about privacy, confidentiality, low-resolution measurements, missing observations, the cost of commercially available data, or minimal motivation for people to measure and share their location data.

To solve this problem, researchers and practitioners can attempt to fix low-quality visit sequences from real people, or they can generate completely synthetic visit sequences. Both approaches lead to the problem of filling gaps in the sequence. In the case of a real visit sequence, with missing parts due to privacy concerns, poor measurements, or dropouts, we have a partially specified sequence. The gaps can be filled with purely the most likely computed visits, or they can be filled with visits that meet some prior background knowledge of where the person went, such as a time-space cube. For instance, the gap may come with only an approximate location, such as somewhere in the vicinity of certain cell tower, which constrains the filled-in visits.

Likewise, for synthetic sequences, we may want to intentionally drop certain visits and replace them with other, loosely specified visits to simulate certain behavior. This is a way to simulate temporarily popular hot spots (e.g. a concert) or travel to new points of interest or newly developed neighborhoods.

For both cases, filling visits in real data or replacing visits in synthetic data, the problem becomes one of replacing gaps with likely visits to complete the sequence, with possible constraints on the filled-in visits. We define this problem of filling gaps as a new challenge that we call ``controlled'' synthetic trajectory generation.

%We define the problem of filling gaps as a new challenge that we call ``controlled” synthetic trajectory generation. 
Gaps in the data present unique challenges, and to the best of our knowledge there are currently no methods designed for this problem. Traditional models for next-location prediction or synthetic generation are not equipped to effectively handle realistically filling gaps in a partially specified sequence. This is challenging for several reasons. First, sequence generation may be subject to certain pre-specified constraints, e.g., location and time of a hot spot being modeled. Second, the number of visits to insert into a gap is unspecified. Finally, the filled-in visits must be specified with not only a realistic visit location (based on an agent's history), but also an accompanying arrival time, visit duration, and travel time that conform to the location choice. For example, if the location choice is a dentist, we will likely not specify a 3 a.m. arrival time and a four-hour visit after an eight-hour drive.

%, e.g. not arriving at a dentist at 3 a.m. for a 4-hour visit. 

%Finally, reasonable travel times to the visit locations must also fit into the gap. 

\begin{figure*}[htbp]
\centering
\begin{subfigure}[t]{0.47\textwidth}
    \includegraphics[width=\textwidth]{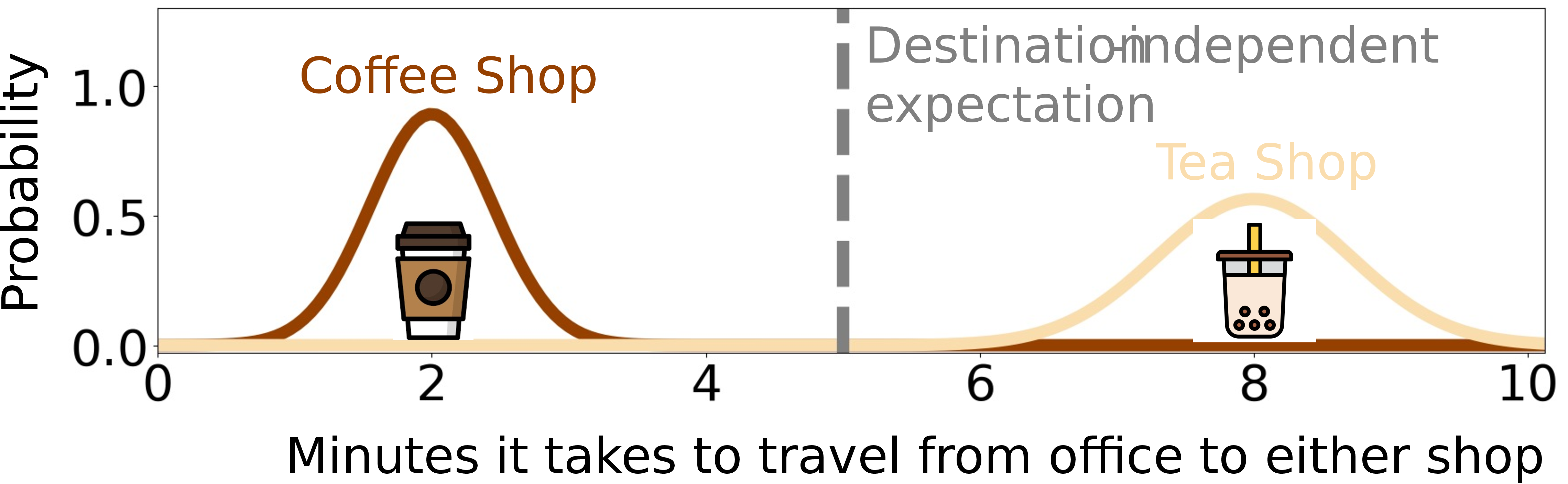}
    \caption{Importance of probabilistic dependence between location and time: An agent at their office will travel to either the coffee shop or tea shop, each with equal probability. It takes eight minutes to reach the tea shop (lighter PDF curve) and two minutes to reach the coffee shop (darker PDF curve). Without knowing the destination, a model might predict an unrealistic five-minute travel time, the average of the two.}
    \label{fig:tea}
\end{subfigure}
\hfill
\begin{subfigure}[t]{0.47\textwidth}
    \includegraphics[width=\textwidth]{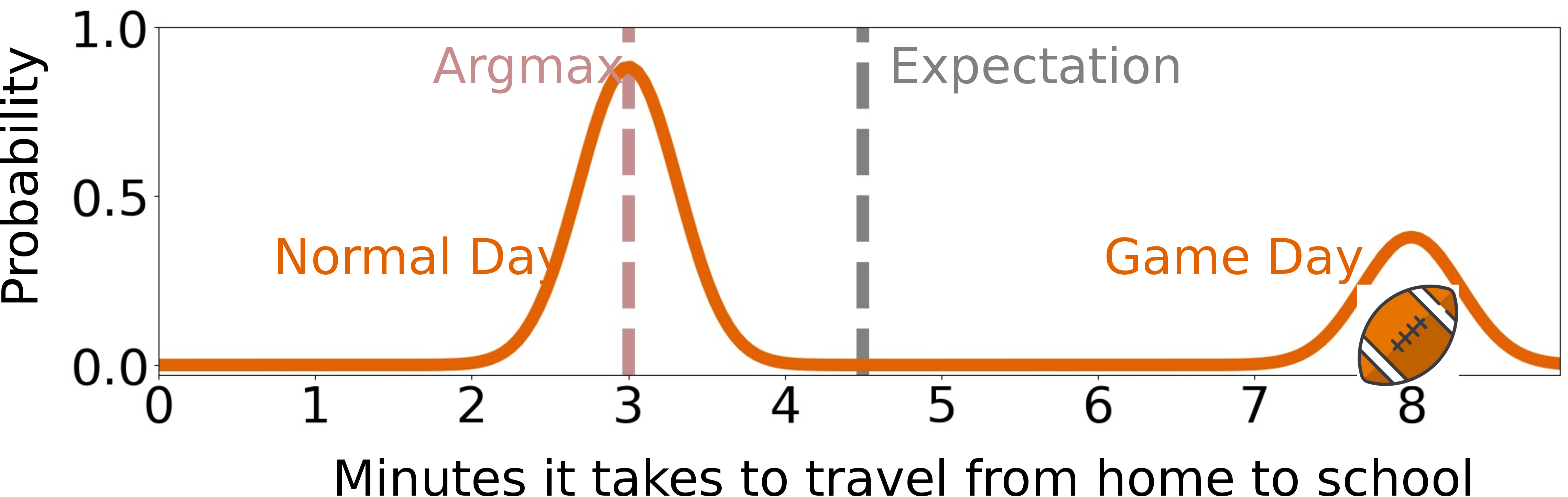}
    \caption{Drawbacks of predicting expected value or most probable point: An agent traveling from home to school faces variable traffic based on whether a sports game is scheduled. Predicting the expected time yields a low probability value, while predicting the most probable timestamp misses all game day scenarios. Additionally, if the predicted arrival time falls outside a gap in a sequence of visits, there's no clear method to adjust it to fit within the gap.}
    \label{fig:gameday}
\end{subfigure}
\caption{Examples motivating (a) spatiotemporal joint probability modeling and (b) Gaussian mixture models, with their corresponding probability density functions (PDFs).}
\label{fig:examples}
\end{figure*}

Existing models for next-location prediction or synthetic trajectory generation lack the necessary mechanisms to constrain the generated sequences of visits. Typically, these models predict only the location of visits \cite{mobtcast, starhit}. While some recent methods have attempted to model both location and time \cite{deepjmt, getnext}, they have significant limitations that affect their performance and the realism in generated trajectories:

\paragraph{Assumption of independence between location and time:} Existing methods frequently treat location and time as independent factors, relying heavily on an independence assumption that fails to hold true in real-world scenarios. For example, suppose an agent leaves their office at lunchtime and is equally likely to visit either a coffee shop or a tea shop. It takes eight minutes to travel from the office to the tea shop, while it takes only two minutes to reach the coffee shop. A model that treats location and time independently might predict the mean travel time (five minutes) regardless of the actual destination, thereby introducing unrealistic artifacts into the generated trajectory. We visualize this example in Figure~\ref{fig:tea}.

\paragraph{Temporal accuracy challenges in single-value time predictions: } Existing methods usually predict a single value of time, such as an expected value derived from regression~\cite{getnext} or the most probable value determined by the argmax of probability~\cite{deepjmt}. This approach can compromise the temporal accuracy of generated visit sequences. For instance, traffic congestion around a school tends to be significantly heavier on game days compared to other days. To accurately predict realistic arrival times while considering such factors, a model should implicitly distinguish between game days and non-game days instead of simply averaging the two possibilities. We illustrate this in Figure~\ref{fig:gameday}. 
\\

These limitations are even more pronounced when the task involves filling gaps within sequences rather than solely predicting the next visit. Existing methods fail to offer effective solutions for either i) generating a realistic sequence of visits with joint spatiotemporal modeling or ii) adequately controlling the output of a regression model while adhering to strict constraints.

To address these issues, we propose TrajGPT, a transformer-based, multi-task, joint spatiotemporal generative model. TrajGPT leverages the transformer architecture to predict locations, while the visit duration and travel time between visits are approximated by taking into account the predicted location. Taking inspiration from recent large language models~\cite{kenton2019bert,gpt3}, TrajGPT poses the problem of controlled trajectory generation as that of text infilling in natural language. By allowing the pre-fixing of specific locations and times within a sequence, TrajGPT can effectively fill in the gaps in a manner that maintains spatiotemporal consistency.

TrajGPT also learns the parameters of a Gaussian mixture to model the distributions of visit duration and travel time between visit locations. The integration of spatial and temporal models is facilitated through a Bayesian probability model, incorporated as a nonparametric joint likelihood loss function. This innovative approach ensures that TrajGPT can fill gaps and generate sequences that are both spatially and temporally consistent. It explicitly avoids the problems illustrated in Figure~\ref{fig:examples} due to its joint probability representation. This capability is crucial for applications requiring precise control over synthetic trajectory generation, such as simulating movement patterns in a partially known scenario.

Our extensive experiments on both public and private datasets highlight the effectiveness of TrajGPT in controlled synthetic visit generation. The results demonstrate that TrajGPT not only excels in filling gaps within sequences but also outperforms competing models in next-location prediction tasks. This superior performance underscores the potential of TrajGPT to advance the field of human mobility modeling by providing a robust and flexible solution for generating controlled synthetic trajectories.\footnote{The code is available at \url{https://github.com/ktxlh/TrajGPT}.}

In summary, the main contributions of this work are:
\begin{enumerate}[topsep=1pt,itemsep=0ex,leftmargin=*]
    \item We introduce the novel problem of ``controlled'' synthetic trajectory generation, which addresses the need to fill gaps in sequences with specific constraints on locations and times.
    \item We propose TrajGPT, a transformer-based, multi-task, joint spatiotemporal generative model that integrates a Gaussian mixture model and a Bayesian probability model to ensure spatiotemporal consistency and accuracy.
    \item We demonstrate the effectiveness of TrajGPT through extensive experiments on both public and private datasets, highlighting its superior performance in controlled synthetic visit generation and next-location prediction tasks compared to existing models.
\end{enumerate}

In the remainder of the paper, we describe related work in Section~\ref{sec:related-work}. Section~\ref{sec:problem_statemet} gives a precise definition of the new problem we solve, and Section~\ref{sec:method} presents our solution in the form of likelihood maximization with probability distributions computed from a transformer model. Section~\ref{sec:evaluation} presents our performance evaluation on trajectory data, showing how our approach is superior to the state-of-the-art alternatives, as well as an ablation study. We conclude in Section~\ref{sec:conclusion}.

\section{Related Work}\label{sec:related-work}
Human mobility data can be categorized into two types: point-based and visit-based. While both are often grouped under the term ``trajectories,'' they differ significantly. Point-based data consists of a sequence of observations from sensors tracking an object's movement, such as GPS signals from a mobile phone \cite{geolife3,mdc-data,reality-data,gonzalez2008understanding} or a car's navigation system \cite{zen-data,tdrive-data,crawdad-data}. This type of data is typically dense, with observations collected at frequent intervals (e.g., every 30 seconds) and includes raw coordinates (latitude and longitude) without any associated semantic information. On the other hand, visit-based data captures the sequence of places an individual visits. Each visit includes details such as location (both latitude/longitude and semantic information like points of interest), arrival time, and departure time. 

For forecasting tasks such as next location prediction \cite{he2024hierarchical,deepstpp,dtmc}, point-based data is rich and easier to predict due to its short and regular inter-point intervals and additional contextual knowledge like road networks \cite{newson2009hidden}. However, this type of sequence is not the focus of our paper.

Our focus is on visit-based data, which is typically sparser and more challenging to predict. This data is often collected through check-ins (e.g., from Foursquare \cite{foursquare_dataset} and Gowalla \cite{gowalla-data}) and usually includes 3-5 visits per person per day with irregular inter-arrival times.
Research on visit-based sequences often centers on the downstream task of predicting the next location \cite{getnext,starhit,deepjmt,mobtcast,zhao2020go,deepmove,stan,stgn,asppa,llmmob}. These locations are typically represented as Points-of-Interest (POIs). Among the papers on POI recommendation, DeepJMT \cite{deepjmt}, MobTCast \cite{mobtcast}, GETNext \cite{getnext}, and STAR-HiT \cite{starhit} are the most relevant to our work as they employ transformers \cite{transformer} as the underlying encoder. While MobTCast and STAR-HiT predict the subsequent POI without considering check-in time, MobTCast integrates various contextual factors, including temporal, semantic, social, and geographical contexts, alongside a consistency loss mechanism. Conversely, STAR-HiT, featuring a hierarchical transformer architecture, employs stacked encoders and subsequence fusion modules to capture multi-granularity spatiotemporal patterns within user check-in sequences, facilitating interpretability.

DeepJMT and GETNext predict location and time independently, although the time spent at a location (e.g., coffee shop vs. gym) and the time between locations are highly dependent on the specific locations. For instance, GETNext employs a deterministic approach to predict a single temporal value, while DeepJMT predicts a temporal distribution during training and predicts the timestamp with the highest probability for inference. While the assumption of independence might not significantly impact the task of next location prediction, it poses a challenge when predicting the arrival time or duration of the next visit, as demonstrated in our experiments (see Section~\ref{sec:evaluation}) using these approaches as baselines. This issue becomes even more pronounced when trying to fill in the visits between known visits, which is the primary focus of our paper.
 
Although no existing work in human mobility modeling directly addresses the task of filling visit gaps, related concepts can be found in language modeling. Following the introduction of the transformer model and large language models (LLMs) \cite{gpt3,t5,xlnet,roberta,unilm,bart}, subsequent research adapted its encoder architecture for masked language modeling (MLM), exemplified by BERT \cite{kenton2019bert}, and its autoregressive decoder for causal modeling, as demonstrated by GPT \cite{gpt}. However, these popular approaches each have their drawbacks when it comes to filling in gaps. BERT can only fill known-length gaps, which is inadequate given the variability in spans of gaps. Conversely, GPT relies solely on the preceding context, lacking the ability to leverage information following a gap in a sequence. Consequently, neither model effectively addresses the challenge of infilling variable-sized gaps constrained by contextual factors. Some previous studies \cite{infilling,blm,glm} have tackled this unknown-length blank infilling problem; thus, we follow the infilling paradigm from natural language processing (NLP) for our specific task of inferring human visit sequences. Nevertheless, this infilling approach cannot be directly applied to our problem due to the absence of a spatial and temporal component, which is crucial in our context where location, visit duration, and inter-arrival timings are important. Therefore, we leverage Space2Vec \cite{space2vec} and Time2Vec \cite{time2vec} for spatiotemporal representation learning and design spatiotemporal joint prediction for controlled synthetic trajectory generation.

Geo-CETRA \cite{geocetra}, a recent study, also addresses the problem of constraint-based trajectory generation, but our work differs in several key ways. In Geo-CETRA, the constraints are defined as spatiotemporal ranges that the synthetic trajectory must satisfy, whereas we define constraints as a set of known visits that the trajectory is required to pass through. As a result, Geo-CETRA focus on identifying realistic visits within the spatiotemporal boundaries, while our approach, inspired by language models, concentrates on filling in the visits between these fixed points. Due to the nature of our discrete constraints, we discretize space into grid cells to construct a ``vocabulary'' for our model, whereas Geo-CETRA operate directly in a continuous spatiotemporal space.

\section{Problem Statement} \label{sec:problem_statemet}
We give a precise definition of our problem here, followed by our solution in Section~\ref{sec:method}.

\subsection{Terminology}
We formally define a \emph{visit} as a tuple $x=(r, t^a, t^d)$, where $r$ represents the location of the visit, such as a region or a Point-of-Interest (POI), $t^a$ represents the arrival time of the visit, and $t^d$ represents the departure time of the visit. A sequence of visits $X$ contains all visits made by a single agent within a time range. We use $P$ to denote the \emph{true} possibility mass function (PMF) or possibility density function (PDF), and $\hat{P}$ to denote the PMF or PDF approximated by TrajGPT. We summarize the notations in Table~\ref{tab:notation}.
\begin{table}[]
    \centering
    \begin{tabular}{c|l}
    \toprule
    % Notation & Description \\
    % \midrule
    $r_i$ & the region where visit $i$ is located, such as a grid cell \\
    $t_{i}^a$ & the arrival time of visit $i$ \\
    $t_{i}^d$ & the departure time of visit $i$ \\
    $x_{i}$ & an attributes tuple $(r_i, t_{i}^a, t_{i}^d)$ of visit $i$ \\
    $X$ & a sequence of contiguous visits $X = [x_1, x_2, ..., x_i]$ \\
    $X'$ & a subsequence of $X$, input for tasks \\
    $P$ & a true possibility mass or density function \\
    $\hat{P}$ & an approximated possibility mass or density function \\
    $H$ & a sequence of visit embeddings \\
    $\Delta t_{i}^{\mathcal{T}}$ & the travel time from visit $i-1$ to $i$ \\
    $\Delta t_{i}^{\mathcal{D}}$ & the duration of visit $i$ \\
    \bottomrule
    \end{tabular}
    \caption{Notable notations used in this article.}
    \label{tab:notation}
\end{table}

\subsection{Controlled Synthetic Trajectory Generation} \label{sec:infillingtask}

The main problem we solve, ``Controlled Synthetic Trajectory Generation,'' is to, given an incomplete sequence of visits $X'$, predict the missing visits within $X'$. Let $X$ be a (complete) sequence of visits. An incomplete sequence of visits, $X'$, refers to any sub-sequence of $X$ that is missing at least one visit. If each contiguous span of missing visits is replaced with a placeholder marker (i.e., a blank), then, the task is to predict the missing visits $\hat{X}$ for each blank, specifying both the temporal ordering of such predicted missing visits and the correspondence of the predicted missing visits to the blanks. Given the predicted missing visits $\hat{X}$ and incomplete visit sequence $X'$, it is trivial to construct the resultant (complete) sequence of visits. If $x$ is the first missing visit within the incomplete visit sequence $X'$, the probability distribution $P(x | X')$ can be rewritten as $P(x | X') = P(r, t^a, t^d | X') = P(r | X') P(t^a | X', r) P(t^d | X', r, t^a)$ according to the chain rule of probability.

\subsection{Next Visit Prediction} \label{sec:nextvisittask}
As a byproduct of solving ``Controlled Synthetic Trajectory Generation,'' we can also solve ``Next Visit Prediction,'' which is a generalized version of the traditional ``Next Location Prediction'' task. Given context $X'$ which consists of a contiguous sequence of visits, we model not only the probability distribution $P(r | X')$ of the next visit's location $r$, but also the probability distribution $P(t^a | X', r)$ of the arrival time $t^a$ of the visit, as well as the probability distribution $P(t^d | X', r, t^a)$ of the departure time $t^d$ of the visit. Therefore, given $X = [x_{1}, ..., x_{i - 1}]$, which is a sequence of consecutive visits, we predict the next visit $x_{i}$, which includes its region $r_{i}$, its arrival time $t_{i}^a$, and its departure time $t_{i}^d$. We make the trivial observation that the Next Visit Prediction task is a special case of the Controlled Synthetic Trajectory Generation task where the only missing visits, $X \setminus X'$, are those occurring in the future, following the last visit in $X'$.

%%%%%%%%%%%%%%%%%%%%%%%%%%%%%%%%%%%%%%%%%%%%%%%%%%%%%%%%%%%%%%%%%%%%%%%%%%%%%%%%
\section{Methodology}\label{sec:method}
In this section, we introduce our solution to the problem of ``Controlled Synthetic Trajectory Generation.'' To utilize techniques from autoregressive sequence modeling, we begin by rearranging each visit sequence, enabling the use of an autoregressive model for the infilling task (Section~\ref{sec:infill}). Subsequently, we design a spatiotemporal autoregressive model that learns its parameters from these rearranged visit sequences (Section~\ref{sec:modeling}).

\subsection{Visit Infilling}\label{sec:infill}

In order to: (1) train the model to be capable of infilling any number of items into each blank, with one or more blanks at any point in the sequence, and  (2) take advantage of the auto-regressive nature of transformers and allow efficient training of the model to do both infilling and next-item prediction, we restructure our sequence data for the infilling task (Section \ref{sec:infillingtask}), following the approach outlined by \citet{infilling}. Each visit sequence $X$ is composed of visits $X = [x_{1}, ..., x_{n}]$, and each visit $x_{i} = (r_{i}, t_{i}^a, t_{i}^d)$ is defined by its region, arrival time, and departure time, respectively. The dataset as it is in its innate form, which consists of many visit sequences, can be used to train a transformer model to predict the next visit, given a partial visit sequence. We then reframe our data for the infilling task by applying the following process to each visit sequence $X = [x_{1}, ..., x_{n}]$.

\begin{figure}[htbp]
  \centering
  \includegraphics[width=0.45\textwidth]{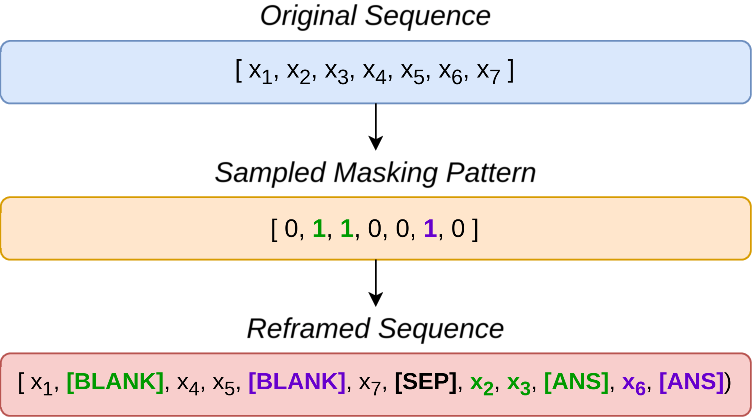}
  \caption{Reframing sequence data for infilling.}\label{fig:infilling_task}
\end{figure}

To rearrange a visit sequence for the infilling task, we first add a special \texttt{SEP} non-visit token to the end of the sequence to denote the end of the original sequence. Then, we sample a Bernoulli distribution for each visit except for the first and last ($x_{i}; i \in \{ 2, 3, ..., n - 1 \}$). Sampling a $1$ means we drop the visit, and a $0$ means we retain the visit. For each contiguous span of visits we dropped, we insert a single \texttt{BLANK} token where the span used to be located within the sequence. Next, for each span we dropped, we append that span and an \texttt{ANS} token to the end of the sequence; the \texttt{ANS} token marks the end of each span. In this way, the reframed sequence contains the partially specified sequence in the first half and the ground truth filled-in visits in the second half. Since this is an infilling task, we never drop the first or last visit. See Figure~\ref{fig:infilling_task} for an example.

% For example, if we start with the abstracted visit sequence (Home, Work, Gym, Home, Work, Restaurant, Home), we may sample a infilling masking pattern as (0, 1, 1, 0, 0, 1, 0), where the first and last element in the masking pattern is necessarily $0$ and not sampled. In that case, we would alter the sequence to become (Home, BLANK, Home, Work, BLANK, Home, SEP, Work, Gym, ANS, Restaurant, ANS).

The aim of the model is to predict the values after the \texttt{SEP} token in order to complete a sequence. This coincides with the prediction of the missing items.
By rearranging each sequence and marking special delimiters (\texttt{BLANK}, \texttt{SEP}, \texttt{ANS}), an autoregressive transformer model can learn to attend to the positions of these special tokens. In doing so, it can start infilling any number of items for the first blank after the \texttt{SEP} token, declaring the first blank to have been completely infilled by predicting an \texttt{ANS} token, and repeating this process for subsequent blanks until the number of \texttt{ANS} tokens matches the number of \texttt{BLANK} tokens. The task of interest is to predict the visits in the reframed sequence after the \texttt{SEP} token. Note that for our training, validation, and test split data, we assume that we have access to complete visit sequences, to which we can apply this reframing process. However, at inference time, we have incomplete visit sequences to be infilled, which constitute the visits before the \texttt{SEP} token.

\subsection{Spatiotemporal Joint Modeling}\label{sec:modeling}
In this section, we discuss the architecture of TrajGPT and explain the process by which it learns the model parameters. We use this architecture and learning process for both Controlled Synthetic Trajectory Generation and Next Visit Prediction. It is important to note that we employ teacher forcing throughout the training phase. For instance, when predicting an arrival time, we use the actual region as input rather than a predicted one.

\subsubsection{Formulation}
We derive a probabilistic model for spatiotemporal autoregressive sequence modeling as follows.
As operationalized in Section~\ref{sec:infillingtask}, to predict the remaining visits $X \setminus X'$ given $X'$, we parameterize a function $\hat{P}$ to approximate the conditional joint probability $P(x \mid X')$, denoting the parameters as $\theta$, and learn it with maximum likelihood estimation (MLE):
\begin{equation}\label{eq:pstar}
\theta^{*} = \text{arg}\max_{\theta} \prod_{x \in X \setminus X'} \hat{P}_{\theta}(x \mid X')
\end{equation}
For simplicity, throughout this article, $X'$ evolves as we add new, inferred visits, and we will omit $\theta$ from our notation going forward.

To achieve the approximation, we first factorize the targeted joint probability using Bayes' Rule:
\begin{align}
\begin{split}
P(x \mid X') 
&= P(r,\ t^a,\ t^d  \mid X') \\
&=  P(r \mid X')\ 
    P(t^a\ \mid X',\ r)\ 
    P(t^d  \mid X',\ r,\ t^a)
\end{split}
\label{eq:bayes}
\end{align}

To approximate these factors, for each visit, we make TrajGPT approximate the distribution of each attribute of $x$ one by one as follows. In other words, TrajGPT predicts region, arrival time, and departure time of a visit sequentially, taking all previous predictions into consideration when making a new prediction.
\begin{enumerate}[leftmargin=*]
    \item Approximate region $P(r \mid X')$
    \item Conditioned on region, approximate arrival time $P(t^a\ \mid X',\ r)$
    \item Conditioned on region and arrival time, approximate departure time $P(t^d  \mid X',\ r,\ t^a)$
\end{enumerate}
We will elaborate on the realization of these steps in the subsequent sections. To define a loss function that encourages TrajGPT to predict the truth, we follow the Maximum Likelihood Estimation (MLE) paradigm and compute the negative log likelihood of predicting the ground truth for each of these approximated distributions, such as $-\log\hat{P}(r_n \mid X')$. Combining these likelihood variables with Equation \ref{eq:pstar} and \ref{eq:bayes}, we obtain this elegant, non-parametric, negative log likelihood loss function for the joint probability\footnote{For special tokens (see Section~\ref{sec:infill}), region loss is replaced with the special token loss, and there is no temporal loss.}:
\begin{equation}
\mathcal{L} = -\sum_{r, t^a, t^d}
    \log\hat{P}(r \mid X')
    + \log\hat{P}(t^a \mid X', r)
    + \log\hat{P}(t^d \mid X', r, t^a)
\end{equation}
where $L$ stands for \emph{loss} and sums over $(r, t^a, t^d) \in\ X \setminus X'$.

\subsubsection{Model Architecture}
\begin{figure*}[htbp]
  \centering
  \includegraphics[width=\textwidth]{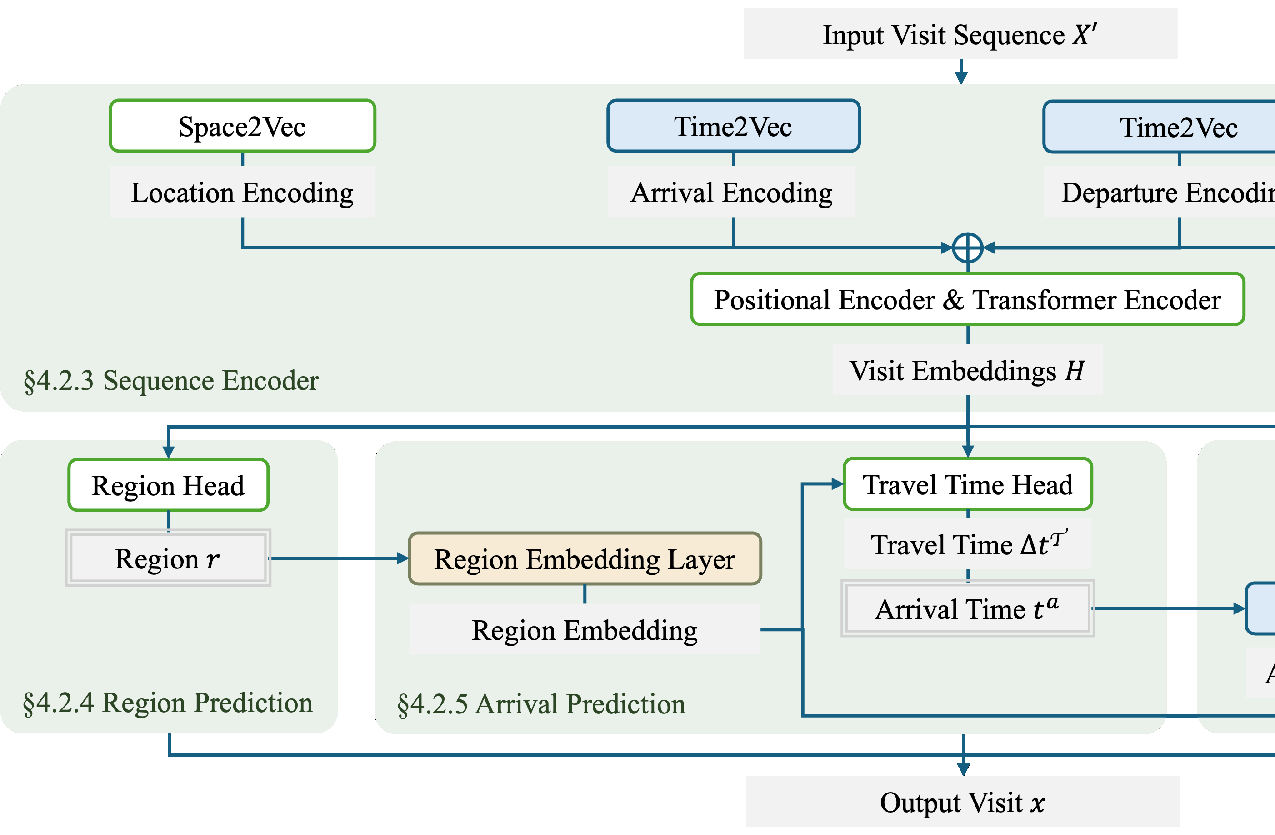}
  \caption{Overview of TrajGPT. Modules that share parameters are colored in the same shade.}\label{fig:arch}
\end{figure*}

We illustrate the architecture of TrajGPT in Figure \ref{fig:arch}. The process begins with fusing the spatiotemporal information in the subsequence of visits \(X'\) using a transformer encoder (Section \ref{sec:seq-encoder}). Following this, the region head module predicts the region \(r\) of the visit (Section \ref{sec:region-pred}) as a discrete probability mass function over possible visit locations. Subsequently, the model embeds and conditions on the predicted region to forecast the travel time of the visit using the travel time head. The travel time is then arithmetically converted to arrival time (Section \ref{sec:arrival-pred}). The arrival time is encoded and fed to the duration head to predict the duration of the visit. Finally, the duration is converted to departure time through arithmetic operations (Section \ref{sec:departure-pred}).

\subsubsection{Sequence Encoder}\label{sec:seq-encoder}
We design a sequence encoder to help TrajGPT understand complex spatiotemporal sequences. For each visit, we use Space2Vec \cite{space2vec} to encode the location $l_i$, known as location encoding, and Time2Vec \cite{time2vec} to encode the arrival and departure times, referred to as arrival and departure time encoding. To guide the model in recognizing region-specific information, such as land use, we embed the region where each visit occurs and make this embedding learnable, referring to it as region embedding. If a visit is a ``special token'' visit, as described in Section \ref{sec:infill}, we use the embedding of the special token instead of a region embedding since this pseudo visit does not contribute spatiotemporal information to the sequence.
We then concatenate the location encoding, arrival time encoding, departure time encoding, and region or special token embedding. This sequence of concatenated embeddings is fed into the positional and transformer encoders proposed by \cite{transformer}. The sequence of outputs from the transformer encoder will be referred to as visit embeddings $H$, which entails an implicit summary of the input sequence.
\begin{equation}
    H\ :=\ \text{TransformerEncoder}(\text{PositionalEncoder}(X'))
\end{equation}

\subsubsection{Region Prediction}\label{sec:region-pred}
We formulate the region prediction task as a classification problem. To predict the region where a visit is located, we feed the visit embeddings $H$ to the region head, which contains another transformer encoder, a linear layer, and a softmax function for creating a proper probability mass function. 
\begin{equation}
    \hat{P}(r_i \mid X')\ :=\ \text{Softmax}(\text{Linear}(\text{TransformerEncoder}(H)))
\end{equation}

\subsubsection{Arrival Time Prediction}\label{sec:arrival-pred}
To account for spatiotemporal dependencies, as shown in Figure~\ref{fig:tea}, we condition our arrival time predictions on the predicted region.

Arithmetically, to predict the arrival time $t_i^a$ of visit $i$, we first derive the travel time $ \Delta t_i^{\mathcal{T}}$ from the preceding visit $i-1$ to $i$, the current one we are predicting.
\begin{equation}
    \Delta t_i^{\mathcal{T}}\ =\ t^a_i\ -\ t_{i-1}^d
\end{equation}

To predict $\Delta t_i^{\mathcal{T}}$, we approximate its potentially complex distribution, as illustrated in Figure~\ref{fig:gameday}, using a Gaussian Mixture Model (GMM). We will show that this approximation effectively models travel time based on visit-to-visit observations in the training data in Section~\ref{sec:evaluation}.
\begin{equation}
\hat{P}(\Delta t_i^{\mathcal{T}} \mid X',\ r_i)\ :=\ P_i^{\mathcal{T}}(\Delta t_i^{\mathcal{T}})
\end{equation}
where $P_{i}^{\mathcal{T}}: \mathbb{R} \rightarrow \mathbb{R}^{+}$ is the probability density function (PDF) of the GMM\footnote{\label{fn:clip-dist}For inference, we clip the distribution by setting the probability of negative values to zero and re-normalizing it.}. To predict the parameters of the GMM, denoted as Param($P_i^{\mathcal{T}}$), we emulate the decoder of transformer \cite{transformer} to enable cross attention between $r_i$ and $H$:\footnote{For brevity, we omit the residual connections and normalization layers in Equation \ref{eq:arrival-pred} and \ref{eq:depart-pred}. For details, please refer to Section 3 of the transformer paper \cite{transformer}.}
\begin{equation}\label{eq:arrival-pred}
\text{Param}(P_i^{\mathcal{T}})
\ :=\ \text{FF}(\text{MHA}(e_i,\ H,\ H))
\end{equation}
where $e_i = \text{Embedding}(r_i)$ is embedded using the same region embedding layer as Section \ref{sec:seq-encoder}; FF denotes feedforward neural networks;\footnote{To ensure the predicted weights and scales of the GMM are always positive, we apply a softplus function and add a small positive value to the output of the feedforward network. Equations are omitted for brevity.} MHA stands for multi-head attention, which projects $\mathbf{r_i}$ to \emph{queries}, and $H$ to \emph{keys} and \emph{values} to perform cross attention:
\begin{align}
\begin{split}
\text{MHA}(e_i, H, H) 
&:= \text{Concat}(\text{head}_1,...,\text{head}_{M}) W^O \\
\text{where}\ \text{head}_j
&:= \text{Attention}(e_iW_j^Q, HW_j^K, HW_j^V)
\end{split}
\end{align}
where \emph{Concat} denotes concatenation of vectors; $W_j^Q; W_j^K, W_j^V$ are parameter matrices.

\subsubsection{Departure Time Prediction}\label{sec:departure-pred}
Similar to how we predict arrival time in Section \ref{sec:arrival-pred}, we first derive duration $\Delta t_i^{\mathcal{D}}$
\begin{equation}
\Delta t_i^{\mathcal{D}}\ =\ t_i^d\ -\ t_i^a
\end{equation}
Then, we use cross attention and a GMM to predict the duration $\Delta t_i^{\mathcal{D}}$ of the visit
\begin{equation}
\hat{P}(\Delta t_i^{\mathcal{D}} \mid X',\ r_i,\ t_i^a)\ := P_i^{\mathcal{D}}(\Delta t_i^{\mathcal{D}})
\end{equation}
where $P_i^{\mathcal{D}}: \mathbb{R} \rightarrow \mathbb{R}^{+}$ is the PDF of the GMM.
Different from arrival time prediction, instead of using only the embedding of $r_i$ as the query for MHA, we first concatenate $e_i = \text{Embedding}(r_i)$ with the encoding of $t^a_i$ as input $c_i$
\begin{equation}
c_i := \text{Concat}(e_i, \text{Time2Vec}(t_i^a))
\end{equation}
where the encoding of $t^a_i$ is generated by the Time2Vec encoder in Section \ref{sec:seq-encoder}. Then, we compute the cross attention between $c_i$ and $H$ to approximate the parameters of the GMM.
\begin{equation}\label{eq:depart-pred}
\text{Param}(P_i^{\mathcal{D}})\ 
:=\ \text{FF}(\text{MHA}(c_i,\ H,\ H))
\end{equation}
\begin{align}
\begin{split}
\text{MHA}(c_i, H, H) 
&:= \text{Concat}(\text{head}_1,...,\text{head}_{M}) W^O \\
\text{where}\ \text{head}_j
&:= \text{Attention}(c_i W_j^Q, HW_j^K, HW_j^V)
\end{split}
\end{align}

In summary, in this section, we described how we train TrajGPT to approximate each probability function of the joint probability in Equation~(\ref{eq:bayes}): $P(r \mid X')\ P(t^a\ \mid X',\ r)\ P(t^d  \mid X',\ r,\ t^a)$.

\subsubsection{Inference}
To conduct inference with a model trained using the above methodology, one can replace teacher forcing with autoregression. In other words, to predict one visit, one can follow these steps:
\begin{enumerate}[leftmargin=*]
    \item Conditioned on $X'$, predict the region of the visit, denoted $\hat{r}$.
    \item Conditioned on $X',\ \hat{r}$, predict the arrival time, denoted $\hat{t}^a$.
    \item Conditioned on $X',\ \hat{r},\ \hat{t}^a$, predict the departure time.
\end{enumerate}
Since the effectiveness of such an autoregressive procedure depends on the choice of a decoding algorithm, such as beam search \cite{beamsearch} or nucleus sampling \cite{nucleussampling}, which is not the focus of this work, we resort to evaluating TrajGPT with teacher forcing in Section~\ref{sec:evaluation}.

%%%%%%%%%%%%%%%%%%%%%%%%%%%%%%%%%%%%%%%%%%%%%%%%%%%%%%%%%%%%%%%%%%%%%%%%%%%%%%%%
\section{Evaluation}\label{sec:evaluation}

\subsection{Experimental Setup}
\subsubsection{Data}
We employed two trajectory datasets, GeoLife and MobilitySim, for our experiments.
GeoLife \cite{geolife3} is a public real-world trajectory dataset based in Beijing, featuring data from 102 agents collected between 2008 and 2009.
To demonstrate the scalability of our approach, we also utilized a private, simulated trajectory dataset, MobilitySim. The dataset contains a realistic simulation of 2,000 agents performing daily activities in San Francisco, over a period of 30 days. The simulation contains second-by-second location of each agent, as they perform recurring daily activities, such as going to school or work, as well as occasional recreation and maintenance activities, such as visits to restaurants, gym, and doctors office. The simulation also incorporates daily and weekly patterns, such as work schedules and days off. 
We summarize the dataset statistics in Table~\ref{tab:data}.

\begin{table*}
\centering
\begin{tabular}{l|rrrrrr}
\toprule
    Dataset & \#agent & \#region & \#visit & \#trajectory* & Avg. \#visit/agent & Avg. \#visit/region\\
\midrule
    GeoLife & 102 & 1,369 & 20,278 & 11,724 & 198.80 & 14.81 \\
    MobilitySim & 2,000 & 3,481 & 191,963 & 6,178 & 95.98 & 55.15 \\
\bottomrule
\end{tabular}
\caption{Data Statistics. \#trajectory denotes the number of trajectories for next visit prediction (see Section~\ref{sec:processing}).}
\label{tab:data}
\end{table*}

\subsubsection{Processing}\label{sec:processing}
To convert point-based trajectories into visit-based sequences, we first identify visits \cite{staypoint}. A visit is operationally defined as a (location, arrival time, departure time) tuple, describing where and when an agent remains stationary for a contiguous period of time. For GeoLife, we identify visits spatially within a 200-meter radius and temporally for a minimum duration of 10 minutes. For MobilitySim, we identify visits for each agent based on a minimum period of 6 minutes during which the agent remains perfectly stationary.
After identifying visits, we form ``regions'' by discretizing the locations using Uber's H3 index \cite{h3}. For GeoLife, we set the Uber H3 Resolution to 7, and for MobilitySim, we set it to 10.
We convert the latitude and longitude of visits to the Universal Transverse Mercator (UTM) coordinate system to ensure the two dimensions of the geographical coordinates are on the same scale (in meters). For timestamps, including arrival and departure times, we subtract the oldest arrival time in each dataset from all other timestamps, converting these time differences into seconds. To prevent exploding gradients during training, we normalize the duration and travel time: the duration is scaled to days, and the travel time is scaled to hours.

For Controlled Synthetic Trajectory Generation (Sections~\ref{sec:infilling-results} and \ref{sec:ablation-results}), we treat each agent's visit sequence as an individual instance and divide the set of agents into training, validation, and test sets in an 8:1:1 ratio. Following a strategy similar to \emph{dynamic masking} in RoBERTa \cite{roberta}, we treat ``masking each visit'' as an independent Bernoulli trial with a 20\% probability. However, after dynamic masking, we replace each contiguous subsequence of masked visits with a \texttt{BLANK}. The model is then tasked with predicting an unknown number of visits for each blank.

For Next Visit Prediction (Section~\ref{sec:next-results}), we followed previous work \cite{getnext,starhit} by using a rolling window to extract instances, sorting them chronologically, and splitting them into training, validation, and test sets in an 8:1:1 ratio. We set the size of the rolling window to 128 visits, following \cite{starhit}.

\subsubsection{Metrics}
We evaluate the models with teacher forcing for these metrics:
\textbf{Acc@k} presents the top-k accuracy for location prediction. Note that we report the evaluation on infilling location predictions, not on the predicting the special \texttt{ANS} token which indicates the model is finished predicting for the corresponding blank. % TrajGPT consistently predicts \texttt{ANS} with near-perfect top-1 accuracy for predicting the \texttt{ANS} token, suggesting that the model can reliably avoid infilling more events than appropriate. %(I think we need to report accuracy for ANS if we want to talk about this)?
\textbf{P$_\pm t$} shows the proportion (for scalar\footnote{For fair comparison with clipped distribution (see Footnote~\ref{fn:clip-dist}), we replace negative predictions with zeros.}) or probability (for distribution) of predictions that fall into the $g \pm t$ minutes interval, where $g$ is the ground truth.

\subsubsection{Baselines}
Since controlled synthetic trajectory generation is a new task we propose, we resort to compare with studies in next POI recommendation. We selected the following state-of-the-art baselines:
\begin{itemize}
    % \item \textbf{MobTCast} \cite{mobtcast}: Transformer for next POI recommendation with social context inferred from check-in data, and an auxiliary next location prediction task.
    \item \textbf{STAR-HiT} \cite{starhit}: Hierarchical transformer for next POI recommendation with subsequence aggregation technique.
    \item \textbf{GETNext} \cite{getnext}: Transformer for next POI recommendation with an auxiliary next check-in time prediction task, assuming next POI and next check-in time are independent.
\end{itemize}

Note that GETNext uses POI category information. As our datasets are generated from raw trajectories, not sequences of POIs, they do not contain such information. Hence, throughout this section, we remove its POI category components.

\subsection{Controlled Synthetic Trajectory Generation} \label{sec:infilling-results}
Since controlled synthetic trajectory generation is a new task we proposed, we aimed to evaluate the effectiveness of TrajGPT compared to existing models. For this purpose, we selected GETNext \cite{getnext}, the state-of-the-art model for human mobility that concurrently models both space and time. We adapted GETNext for the visit infilling task, naming it GETNext*, by incorporating special-token visits into the input sequences, as detailed in Section~\ref{tab:infilling}, and adding an additional departure time head with the same architecture as its original arrival time head.

As shown in Table~\ref{tab:infilling}, TrajGPT maintains similar accuracy in region prediction while achieving significantly higher accuracy in arrival and departure time prediction. This outcome is expected because, although GETNext models both the next point of interest (POI) and the next check-in time, its primary focus is on next POI prediction. In fact, GETNext does not report metrics for temporal prediction, which explains the inferior accuracy in its temporal predictions compared to TrajGPT.
\begin{table*}
\centering
\begin{tabular}{l|l|cccc|ccc|ccc}
\toprule
\multirow{2}{3.5em}{Dataset} & \multirow{2}{5em}{Method} & \multicolumn{4}{c|}{Region} & \multicolumn{3}{c|}{Arrival Time} & \multicolumn{3}{c}{Departure Time} \\ 
&& Acc@1 & Acc@5 & Acc@10 & Acc@20 & P$_{\pm5}$ & P$_{\pm10}$ & P$_{\pm20}$ & P$_{\pm5}$ & P$_{\pm10}$ & P$_{\pm20}$ \\
\midrule
\multirow{2}{3.5em}{GeoLife} 
% 4 layers, 2 heads, lr=0.0001
& GETNext* & 7.64 & 36.87 & \textbf{46.83} & 51.87 & 0.16 & 0.38 & 0.73 & 1.41 & 2.87 & 7.91 \\
% 2 layers, 8 heads, lr=0.0001
& TrajGPT & \textbf{19.87} & \textbf{40.57} & 44.78 & \textbf{52.21} & \textbf{65.72} & \textbf{75.53} & \textbf{85.31} & \textbf{36.08} & \textbf{50.81} & \textbf{64.23} \\ 
\bottomrule
\end{tabular}
\caption{Comparison of TrajGPT for the infilling task with GETNext*, which we adapted from GETNext, showing the effectiveness of TrajGPT. The best results are highlighted in \textbf{bold}.}  % with statistical significance (Mann-Whitney U rank test; p<0.01)
\label{tab:infilling}
\end{table*}

\subsection{Next Visit Prediction} \label{sec:next-results}
To ensure a fair comparison with existing approaches, we also adapted TrajGPT to predict the next visit rather than filling in gaps. This modification aligns the task more closely with next POI recommendation, which the baseline models are specifically designed for. The results of this comparison are presented in Table~\ref{tab:next-prediction}. 
Certain cells within the table intentionally remain unpopulated due to the inherent characteristics of STAR-HiT and GETNext: STAR-HiT is not designed to predict timestamps; GETNext predicts only one timestamp for each visit, and we opt to forecast the arrival time.

In the domain of next visit prediction, TrajGPT exhibits notable superiority over GETNext in temporal forecasting, with minimal adverse effects on its region prediction performance. Notably, TrajGPT achieves this without relying on the supplementary trajectory flow map and transition attention map proposed by GETNext. Furthermore, both GETNext and TrajGPT significantly outperform STAR-HiT, this suggests that learning a multi-task, spatiotemporal model, might help predict locations better.

The use of teacher forcing ensures that TrajGPT has access to the actual region when predicting arrival times, whereas GETNext, by design, lacks this advantage as it predicts both region and arrival time simultaneously and independently. 
To peek into the potential of TrajGPT during inference without teacher forcing, imagine the worst case: If the predicted, most-probable region were incorrect, the arrival time accuracy would be zero. In this case, we can multiply the arrival time accuracy of TrajGPT with its Acc@1 of region prediction. This minimum threshold of its arrival time accuracy will still surpass that of GETNext by a significant margin.
\begin{table*}
    \centering
    \begin{tabular}{l|l|cccc|ccc|ccc}
\toprule
\multirow{2}{5em}{Dataset} & \multirow{2}{5em}{Method} & \multicolumn{4}{c|}{Region} & \multicolumn{3}{c|}{Arrival Time} & \multicolumn{3}{c}{Departure Time} \\
&& Acc@1 & Acc@5 & Acc@10 & Acc@20 & P$_{\pm5}$ & P$_{\pm10}$ & P$_{\pm20}$ & P$_{\pm5}$ & P$_{\pm10}$ & P$_{\pm20}$ \\
\midrule
    \multirow{3}{3.5em}{GeoLife} 
    % & MobTCast &&&&&-&-&-&-&-&-\\
    & STAR-HiT &17.92&40.78&48.98&56.40&-&-&-&-&-&-\\
    & GETNext &\textbf{38.74} &\textbf{66.38} &\textbf{72.78} &77.13 &2.39 &4.27 &8.62&-&-&-\\
    & TrajGPT & 35.06 & 62.08 & 70.77 & \textbf{77.63} & \textbf{64.28} & \textbf{71.70} & \textbf{80.02} & \textbf{35.01} & \textbf{48.40} & \textbf{60.82} \\
    \midrule
    \multirow{3}{3.5em}{MobilitySim} 
    % & MobTCast &&&&&-&-&-&-&-&-\\
    & STAR-HiT &42.79&62.87&70.25&75.74&-&-&-&-&-&-\\
    & GETNext &51.46 &\textbf{80.91} &91.59 &94.50 &1.29 &2.91 &5.50&-&-&-\\
    & TrajGPT &\textbf{54.53}& 80.26& \textbf{92.88}& \textbf{94.66}& \textbf{89.33}& \textbf{94.01}& \textbf{98.07}& \textbf{52.57}& \textbf{62.05}& \textbf{71.20} \\
    \bottomrule
    \end{tabular}
    \caption{Comparison of TrajGPT with baseline models for the next visit prediction task. The best results are highlighted in \textbf{bold}.}
    \label{tab:next-prediction}
\end{table*}

\subsection{Ablation Study} \label{sec:ablation-results}
To demonstrate the effectiveness of the key components of TrajGPT, we conducted an ablation study for the infilling task, with results shown in Table~\ref{tab:ablation}. 

\begin{itemize}[leftmargin=*]
    \item The \textbf{TrajGPT w Independence} variant predicts the region, arrival time, and departure time independently, reflecting the spatio-temporal-independence assumption made by DeepJMT \cite{deepjmt} and GETNext \cite{getnext}. This assumption is expressed as:
    \begin{equation}
        P(r_i, t_i^a, t_i^d \mid X') = 
        P(r_i \mid X') P(t_i^a \mid X') P(t_i^d \mid X')
    \end{equation}
    \item The \textbf{TrajGPT w Regression} variant replaces the GMM used for predicting arrival and departure times (Sections \ref{sec:arrival-pred} and \ref{sec:departure-pred}) with a regression head, mimicking the approach used by GETNext \cite{getnext}.
\end{itemize}

The results demonstrate that TrajGPT significantly outperforms both variants in predicting arrival and departure times, while also maintaining exceptional accuracy in regional predictions. This underscores the effectiveness of the proposed spatiotemporal modeling approach.

Replacing joint modeling with independent modeling greatly reduces the accuracy of departure time predictions. This suggests that the duration of a visit, which influences the departure time, varies significantly depending on the visit's location (i.e. the region).

Replacing GMM with regression significantly weakens the model's ability to predict time, highlighting the importance of learning a time distribution rather than relying on a single point estimate. The accuracy drop is particularly pronounced for departure time predictions, suggesting that predicting a point is even less suitable for duration than for travel time.

The performance drop appears more pronounced for temporal predictions than for region predictions. This may be because each variant \emph{directly} impacts temporal predictions by altering either the input (TrajGPT w Independence) or the output (TrajGPT w Regression), whereas region predictions are only \emph{indirectly} affected through the combined influence of the loss function and optimization process.

In summary, the ablation study demonstrates that each of the two key components of TrajGPT, including joint modeling and GMM-based temporal distribution learning, plays a crucial role in achieving high prediction accuracy. Removing or replacing these components leads to a substantial decrease in performance, further confirming their necessity in capturing the complex spatiotemporal patterns of human mobility trajectory data.

\begin{table*}
\centering
\begin{tabular}{l|l|cccc|ccc|ccc}
\toprule
\multirow{2}{5em}{Dataset} & \multirow{2}{5em}{Method} & \multicolumn{4}{c|}{Region} & \multicolumn{3}{c|}{Arrival Time} & \multicolumn{3}{c}{Departure Time} \\
&& Acc@1 & Acc@5 & Acc@10 & Acc@20 & P$_{\pm5}$ & P$_{\pm10}$ & P$_{\pm20}$ & P$_{\pm5}$ & P$_{\pm10}$ & P$_{\pm20}$ \\
\midrule
\multirow{3}{3.5em}{MobilitySim} 
    & TrajGPT w Independence &39.71 & 79.52 & 83.21 & 85.80 & 68.77 & 81.91 & 90.54 & 31.85 & 40.48 & 46.33\\
    & TrajGPT w Regression &43.61 & 80.87 & \textbf{86.31} & \textbf{90.40} & 31.49 & 56.91 & 78.16 & 1.53 & 3.21 & 6.42\\
    & TrajGPT & \textbf{44.15} & \textbf{81.04} & 86.11 & 89.71 & \textbf{73.71} & \textbf{84.40} & \textbf{91.67} & \textbf{42.65} & \textbf{50.45} & \textbf{57.70}\\
\bottomrule
\end{tabular}
\caption{Comparison of TrajGPT with its variants for the infilling task, demonstrating the effectiveness of TrajGPT's design. The best results are highlighted in \textbf{bold}.}  %  with statistical significance (Mann-Whitney U rank test; p<0.05) 
\label{tab:ablation}
\end{table*}

% \section{Dicussion}

\section{Conclusion} \label{sec:conclusion}

In this paper we introduced the novel problem of ``controlled'' synthetic trajectory generation, addressing the need to fill gaps in visit sequences with specific constraints on locations and times. Filling gaps is useful for imputing missing data and for generating synthetic visit sequences that have some preordained visits. The task is challenging because the filled-in visits, along with travel times, must fill the gap exactly, and the visit locations and durations must be realistic. 

As a solution we presented TrajGPT, a transformer-based, multi-task, joint spatiotemporal generative model. TrajGPT leverages the transformer architecture to predict locations while separately approximating the visit duration and travel time between visits using a Gaussian mixture model. This innovative approach ensures that TrajGPT can generate sequences that are both spatially and temporally realistic by adhering to the statistical dependencies between visit locations, visit durations, and travel times.

We validated our approach on a public and private dataset, comparing against state-of-the-art methods for predicting next locations. We observed that TrajGPT not only demonstrates proficiency in gap filling but also surpasses competing methods in predicting the next visit. On average, TrajGPT achieves a remarkable 26-fold enhancement in temporal prediction accuracy while preserving over 98\% of the spatial accuracy achieved by state-of-the-art approach.

% For example we beat SOTA by an average of x% or up to y% in metrics A and something similar for filling the gaps accuracy....

\begin{acks}
Research supported by the Intelligence Advanced Research
Projects Activity (IARPA) via the Department
of Interior/Interior Business Center (DOI/IBC) contract
number 140D0423C0033. The U.S. Government is authorized
to reproduce and distribute reprints for Governmental
purposes, notwithstanding any copyright annotation
thereon. Disclaimer: The views and conclusions contained
herein are those of the authors and should not
be interpreted as necessarily representing the official
policies or endorsements, either expressed or implied, of
IARPA or the U.S. Government.
This research project has benefited from the Microsoft Accelerate Foundation Models Research (AFMR) grant program through which leading foundation models hosted by Microsoft Azure along with access to Azure credits were provided to conduct the research.
\end{acks}

\bibliographystyle{ACM-Reference-Format}
\bibliography{base}

\appendix 
% \newpage
\section{Experimental Settings}\label{sec:hyperparameters}
For all experiments with TrajGPT, the following settings remain consistent for both GeoLife and MobilitySim:
We implement TrajGPT in PyTorch and train it with an AMD EPYC 7V13 64-core CPU and an NVIDIA A100 80GB GPU.
The number of scales for Space2Vec is 64. The largest scale of Space2Vec is set to the diameter of the region of interest, and the smallest scale is set to 1 meter. The dropout in the transformer is set to 0.1. The epsilon of layer normalization for the transformer is set to 1e-5. The learning rate is set to 1e-4. The patience for early stopping is set to 10 epochs. The random seed is set to 0.

We determine the rest of the hyperparameters of TrajGPT through grid search, using the validation loss as the selection criterion.
For experiments on GeoLife, we set the number of layers for all transformer encoders to 2, the number of attention heads for all multi-head attention modules to 8, and the feedforward dimension to 32. GMM contains 3 components. Region and special token embeddings are each 32 dimensions. Each training batch contains 64 instances.
For experiments on MobilitySim, we utilize 4-layer transformer encoders with a feedforward dimension of 256. The number of heads is 2 for all multi-head attention modules. GMM contains 5 components. Embedding size is 64. The batch size is 128.

For both GETNext and STAR-HiT, we employ the implementations from their respective repositories, which are linked to in their papers, and set all hyperparameters according to the papers.

% Hardware, hyperparameters used, ...

\end{document}